\documentclass[10pt,twocolumn,letterpaper]{article}

\usepackage{iccv}
\usepackage{times}
\usepackage{epsfig}
\usepackage{graphicx}
\usepackage{amsmath}
\usepackage{amssymb}
\usepackage[T1]{fontenc}    
\usepackage{url}            
\usepackage{booktabs}       
\usepackage{amsfonts}       
\usepackage{nicefrac}       
\usepackage{microtype}      
\usepackage{graphicx}
\usepackage{color}
\usepackage{adjustbox}
\usepackage{algorithm}
\usepackage{algorithmic}

\usepackage{multirow}

\usepackage{placeins}

\usepackage{threeparttable}
\usepackage{tablefootnote}
\usepackage[normalem]{ulem}
\usepackage{authblk}

\usepackage[toc,page]{appendix}

\usepackage[pagebackref=true,breaklinks=true,letterpaper=true,colorlinks,bookmarks=false]{hyperref}

\definecolor{amber(sae/ece)}{rgb}{1.0, 0.49, 0.0}
\newcommand{\shortrightarrow}[1][3pt]{\mathrel{%
   \hbox{\rule[\dimexpr\fontdimen22\textfont2-.2pt\relax]{#1}{.4pt}}%
   \mkern-4mu\hbox{\usefont{U}{lasy}{m}{n}\symbol{41}}}}

\newcommand{\rob}[1]{{\color{blue}#1}}

\DeclareMathAlphabet{\mathcal}{OMS}{cmsy}{m}{n}

\iccvfinalcopy 

\def\iccvPaperID{4222} 

\ificcvfinal\fi

\begin{document}

\title{Adversarial Robustness for Unsupervised Domain Adaptation}
\author{Muhammad Awais \textsuperscript{\rm 1, 2}\thanks{This work was carried out at Huawei Noah’s Ark Lab. The webpage for the project is available at: \href{https://awaisrauf.github.io/robust_uda}{awaisrauf.github.io/robust\_uda}
},
Fengwei Zhou \textsuperscript{\rm 1},
Hang Xu \textsuperscript{\rm 1},
Lanqing Hong \textsuperscript{\rm 1},\\
Ping Luo \textsuperscript{\rm 3},
Sung-Ho Bae \textsuperscript{\rm 2}\thanks{Corresponding Author},
Zhenguo Li 
}

\affil[1]{Huawei Noah’s Ark Lab}
\affil[2]{Dept. of Computer Science, Kyung-Hee University, South Korea}
\affil[3]{Dept. of Computer Science, The University of Hong Kong}
\affil[ ]{\normalsize{\texttt{awais@khu.ac.kr}, \texttt{\{zhoufengwei, xu.hang, honglanqing\}@huawei.com}, \texttt{pluo@cs.hku.hk}, \texttt{shbae@khu.ac.kr}, \texttt{li.zhenguo@huawei.com}}
\vspace{-1.5em}
}

\maketitle
\ificcvfinal\thispagestyle{empty}\fi

\begin{abstract}
\noindent Extensive Unsupervised Domain Adaptation (UDA) studies have shown great success in practice by learning transferable representations across a labeled source domain and an unlabeled target domain with deep models. However, previous works focus on improving the generalization ability of UDA models on clean examples without considering the adversarial robustness, which is crucial in real-world applications. Conventional adversarial training methods are not suitable for the adversarial robustness on the unlabeled target domain of UDA since they train models with adversarial examples generated by the supervised loss function. In this work, we leverage intermediate representations learned by multiple robust ImageNet models to improve the robustness of UDA models. Our method works by aligning the features of the UDA model with the robust features learned by ImageNet pre-trained models along with domain adaptation training. It utilizes both labeled and unlabeled domains and instills robustness without any adversarial intervention or label requirement during domain adaptation training.  Experimental results show that our method significantly improves adversarial robustness compared to the baseline while keeping clean accuracy on various UDA benchmarks.
\end{abstract}

\section{Introduction}

\noindent Transferring knowledge from a labeled source domain to an unlabeled target domain is desirable in many real-world applications. However, deep learning models do not perform well in the presence of such domain shifts. For example, a model trained on synthetic data may fail to generalize well on real-world data. Unsupervised Domain Adaptation (UDA) seeks to solve this problem by learning domain-invariant features. Recent UDA methods harness transferable features learned by deep models pre-trained on large datasets like ImageNet~\cite{ganin2016domain,  he2016deep, long2017deep, long2018conditional, zhang2019bridging, liu2019transferable, tang2020unsupervised, gu2020spherical, jiang2020implicit, dalib}. However, a large body of work has shown that these deep models are vulnerable to small adversarial changes in the input that can easily fool the trained models~\cite{biggio2013evasion, szegedy2013intriguing, goodfellow2014explaining, carlini2017towards}. The widespread use of these models in sensitive applications requires them to be robust against these changes. 

\begin{figure*}[t]
    \centering
    \includegraphics[width=1\textwidth]{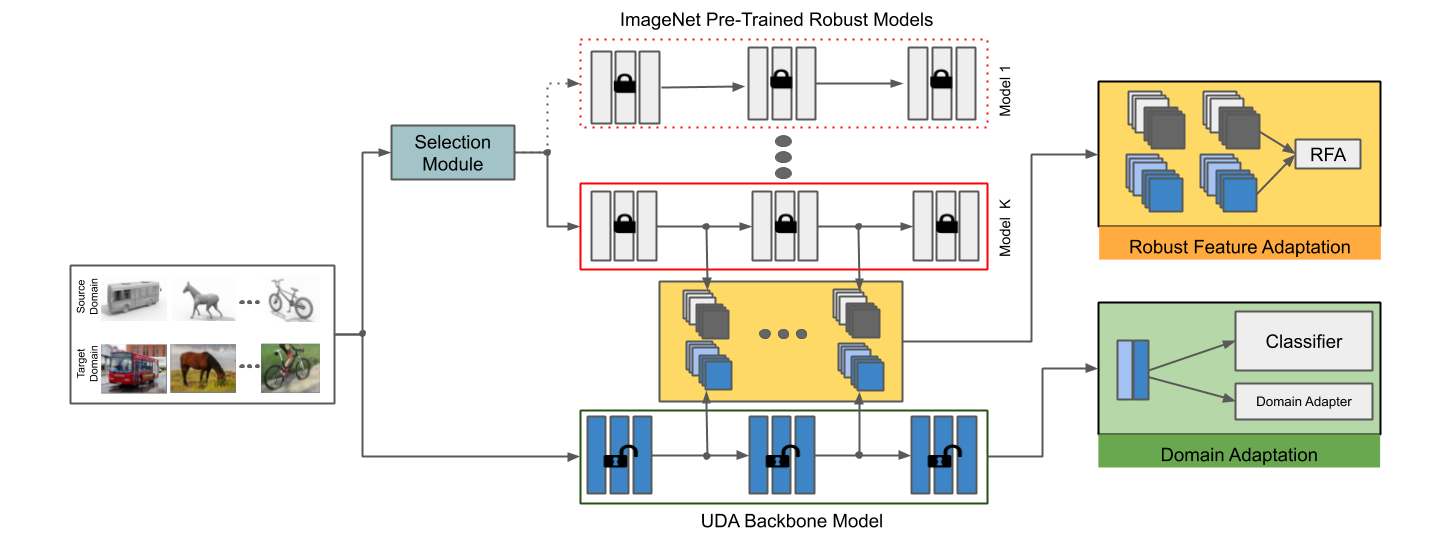}
    \caption{An overview of the proposed method. Source and target images are passed through the backbone model and robust teachers to get features at different blocks. The intermediate features are transferred to the robust feature adaptation (RFA) module, which adapts the robustness. The output of the backbone model goes through the domain adaptation module, which utilizes an unsupervised domain adaption algorithm. The parameters of the UDA feature extractor are updated to minimize both domain adaptation and robust feature adaptation loss. Light colors show the features extracted for source domain inputs and dark colors for target domain inputs.}
    \label{fig:block_rfa}
\end{figure*}

\noindent Significant attention has been devoted to counter adversarial examples, and many defense methods have been devised~\cite{goodfellow2014explaining, guo2017countering,tramer2017ensemble,  madry2017towards, buckman2018thermometer, liao2018defense, ross2018improving, shafahi2019adversarial, tramer2019adversarial, wong2020fast}. Supervised adversarial training is among the most successful approaches~\cite{madry2017towards}. It is based on the simple idea of training a model on adversarial examples. It utilizes min-max optimization where adversarial examples are first generated by iterative maximization of the loss, and the model is then trained on these examples. However, the generation of these adversarial examples requires labels and adversarial training implicitly assumes inputs from a single domain. These issues limit the applicability of adversarial training in UDA. 

\noindent In this paper, we propose a simple, unsupervised, and domain agnostic method for robustness in UDA. It does not require labels and utilizes data from both domains, making it feasible for UDA. Our work is motivated by the recent line of work on transferability of robustness~\cite{goldblum2019adversarially, chan2020thinks}, and observation that adversarially trained models learn "fundamentally different" features from normally trained counterparts~\cite{tsipras2018robustness, ilyas2019adversarial, santurkar2019image}. The first line of work has demonstrated the transferability of adversarial robustness from a pre-trained robust model. The authors in~\cite{hendrycks2019pretraining, shafahi2019adversarially} show that adversarially pre-trained models can improve robustness in transfer learning; \cite{goldblum2019adversarially} shows that adversarial robustness can be distilled by matching softened labels produced by robust pre-trained models; \cite{chan2020thinks} shows that robustness can be distilled by matching input gradients of robust models to those of a non-robust model. These works focus on cutting the computational cost of adversarial training for single domain classification and require labeled data.

\noindent Our proposed method, Robust Feature Adaptation (RFA), embeds the adaptation of robustness in the domain adaptation training by leveraging the feature space of robust pre-trained models. RFA uses ImageNet adversarially pre-trained models to extract robust features for inputs of source and target domains. It then instills robustness in UDA's feature extractor by minimizing its discrepancy with robust features. RFA enables the model to learn both domain invariant and robust features.

\noindent Unlike previous works on transferability, our method does not require labeled data as it only uses intermediate features of the robust models and a label-free distance measure between the feature spaces of the two models. Similarly, RFA does not require any adversarial intervention during the domain adaptation training as it does not generate adversarial examples. These characteristics make it possible to harnesses both labeled source and unlabeled target domains. Moreover, the RFA is a plug-in method that can be used with any UDA method to enhance its robustness. It only requires adversarially pre-trained models similar to the UDA methods that need normally pre-trained models. Our experiments show that RFA can equip UDA models with high adversarial robustness while keeping good generalization ability.
Our contributions can be summarized as follows:

\begin{itemize}
    \item We propose a plug-in method that aligns the features of a UDA model with the robust features of multiple adversarially pre-trained ImageNet models. This way, it instills robustness in UDA models without adversarial intervention or label requirement.
    \item To the best of our knowledge, we are the first to show that the adversarial robustness for a target task can be distilled from intermediate representations of robust models adversarially trained on a different task without any fine-tuning.
    \item Comprehensive experimental results show that our method consistently improves the robustness of various UDA algorithms on widely-used benchmark datasets. For instance, it improves the adversarial robustness from 0\% to 43.49\% while maintaining the clean accuracy for CDAN as the UDA algorithm on challenging simulation-to-real adaptation task of the VisDA-2017 dataset.
\end{itemize}


\begin{table*}
\centering
\begin{adjustbox}{max width=1\textwidth, center}
\begin{threeparttable}
\begin{tabular}{lccccccccccccc}
    \toprule
    
    \multirow{1}{*}{{Dataset}}    & \multicolumn{1}{r}{{Robust PT}} & \multicolumn{2}{c}{{Source-only}}  & \multicolumn{2}{c}{DANN~\cite{ganin2016domain}}    & \multicolumn{2}{c}{DAN~\cite{long2015learning}}              & \multicolumn{2}{c}{CDAN~\cite{long2018conditional} }          & \multicolumn{2}{c}{JAN~\cite{long2017deep}}             & \multicolumn{2}{c}{MDD~\cite{zhang2019bridging} }            \\
    \hline
    &  &Acc. & Rob. &Acc. & Rob.&Acc. & Rob.&Acc. & Rob.&Acc. & Rob. &Acc. & Rob.\\
    \hline
    \multirow{2}{*}{VisDA-17}  & $\times$           & 43.05 &  0        & 71.34 & 0       & 61.79 & 0.01    & 74.23 & 0       & 63.70 & 0       & 72.20 & 4.03    \\
                               & $\checkmark$       & 25.67 & 6.64      & 65.79 & 38.21   & 42.24 & 22.11   & 68.00 & 41.67   & 55.08 & 32.15   & 67.72 & 39.50   \\
    \hline
    \multirow{2}{*}{Office-31} & $\times$           & 77.80 & 0.02      & 85.79 & 0        & 81.72 & 0       & 86.90 & 0      & 85.68 & 0      & 88.31 & 1.70    \\
                               & $\checkmark$       & 69.51 & {41.11}   & 77.30 & 62.38      & 73.71 & 42.29     & 79.67 & 65.53   & 75.12 & 60.24    & 80.72 & 67.54  \\

    \hline
     \multirow{2}{*}{Office-Home} & $\times$        & 58.29 & 0.06    & 63.39 & 0.05     & 59.64 & 0.23    & 67.03 & 0.04   & 64.61 & 0.07   & 67.91 & 5.81  \\
                                  & $\checkmark$    & 53.89 & {31.46}     & 58.10 & {37.25}        & 55.18 & {24.21}       & 63.04 & {43.81}   & 60.74 & 33.09 & 63.30 & 43.42  \\

    \bottomrule  
\end{tabular}
\begin{tablenotes}
	\item \noindent $\times$: Normally Pre-Trained Model, $\checkmark$: Adversarially Pre-Trained Model, PT: Pre-Training.
\end{tablenotes}
\end{threeparttable}
\end{adjustbox}

\caption{Can Robust Pre-Training (PT) instill robustness in unsupervised domain adaptation setting? Comparison between normally and adversarially pre-trained models for clean accuracy and adversarial robustness (\%) with six UDA algorithms. Adversarial pre-training improves adversarial robustness but also causes a drop in clean accuracy.  
}
\label{tab:pretraining_robustness}
\end{table*}

\section{Background}
\subsection{Related Work}
\noindent \textbf{Unsupervised Domain Adaptation. } Most of the unsupervised domain adaptation methods are motivated by the theoretical results in~\cite{ben2006analysis, ben2010theory}. These results suggested learning representations invariant across domains. In deep learning, this is often achieved by min-max training where a pre-trained deep neural network is fine-tuned such that not only does it minimize the loss on labeled data from the source domain but also fool a discriminator. This discriminator is simultaneously trained to distinguish between source and target domains~\cite{ganin2016domain}. In recent works, it has also been shown that large models, pre-trained on large-scale datasets such as ImageNet, improve unsupervised domain adaptation~\cite{long2015learning, ganin2016domain,  he2016deep, long2017deep, long2018conditional, zhang2019bridging, liu2019transferable, tang2020unsupervised, gu2020spherical, jiang2020implicit}. Several unsupervised domain adaptation algorithms have been proposed that leverage pre-trained models~\cite {long2015learning, long2018conditional, zhang2019bridging, liu2019transferable}. However, these works do not consider robustness. Our work is complementary to these works as it improves the robustness of these methods.

\noindent \textbf{Adversarial Training and Robust Features.} Adversarial attacks are considered security risk~\cite{biggio2013evasion, szegedy2013intriguing, goodfellow2014explaining, carlini2017towards}. Numerous methods have been proposed to defend against such examples~\cite{guo2017countering,tramer2017ensemble, madry2017towards, buckman2018thermometer, liao2018defense, ross2018improving, shafahi2019adversarial, tramer2019adversarial, wong2020fast, awais2020towards}. Adversarial training -- the most effective defense mechanism -- is devised to defend against $\ell_p$ bounded adversarial perturbations~\cite{goodfellow2014explaining, madry2017towards} in the inputs. However, adversarial training requires labels and therefore is not suitable for UDA training. In another direction, recent work has also shown that adversarially trained models learn ``fundamentally different'' representations~\cite{tsipras2018robustness, ilyas2019adversarial, engstrom2019adversarial, zhu2021towards}. Our work is motivated by this observation, and we proposed an algorithm to leverage these robust features.

\noindent \textbf{Knowledge and Robustness Transfer.}
The main purpose of knowledge distillation is to decrease the size of a large model. It works by distilling the knowledge of a big pre-trained teacher model to a compact \textit{randomly initialized} student model for the same dataset~\cite{hinton2015distilling}. Many different settings have been explored to achieve this objective \cite{romero2014fitnets, yim2017gift, zagoruyko2016paying, tung2019similarity}. Our work is different from these works as we want only to adapt robustness from the teacher without labels while also learning domain invariant features that perform well on two domains.

\noindent Our work is motivated by~\cite{goldblum2019adversarially, chan2020thinks, hendrycks2019pretraining, shafahi2019adversarially} that showed transferability of robustness. However, these methods are primarily motivated to decrease the computational cost of adversarial training and require labels. In~\cite{goldblum2019adversarially}, the authors showed that the robustness can be distilled from a large pre-trained model (e.g., ResNet) to a compact model (e.g., MobileNet) by utilizing soft class scores produced by the teacher model. Compared to the work in~\cite{goldblum2019adversarially}, our method distills robustness from the intermediate representations only. Furthermore, the distillation is performed from teachers trained on one task (i.e., supervised classification) to a student needed to be trained on another task (i.e., unsupervised domain adaptation), which has not been explored previously. In~\cite{chan2020thinks}, the distillation is performed by matching the gradient of the teacher and student. This method requires fine-tuning on target tasks, back-propagation to get gradients, and discriminator-based learning. Compared to~\cite{chan2020thinks}, our proposed method does not require any fine-tuning, and it adapts robust features from pre-trained models without requiring any extra back-propagation. Moreover, both of these distillation methods require labels and are designed for single-domain training. 

\subsection{Preliminaries}
\label{sec:preliminaries}
\noindent Unsupervised Domain Adaptation aims to improve generalization on target domain by reducing domain discrepancy between source and target. Formally, we are given labelled data in the source domain $D_s = \{(x_i^s, y_i^s)\}_{i=1}^{n_s} \sim P$ and unlabeled data in the target domain $D_t = \{x_j^t\}_{j=1}^{n_t} \sim Q$, where $P\neq Q$. Most unsupervised domain adaptation methods fine-tune a pre-trained backbone model $f(x; \theta)$ and train a classifier $C(f(x; \theta); \psi)$ on top of it. The training is done in such a way that it reduces error on the labeled source domain as well as learning features that are invariant in both source and target domains.  

\noindent Adversarial examples~\cite{szegedy2013intriguing, goodfellow2014explaining} are bounded and imperceptible perturbations in the input images that change the normal behavior of neural networks. Thus, the adversarial robustness of a model is its invariance to such small $\ell_p$ bounded perturbation in the input. To achieve this robustness, adversarial examples are created by maximizing the loss, and then it is minimized to train the model~\cite{madry2017towards}:
\[
\min_{\theta, \psi} \mathbb{E}_{(x, y) \sim D} \bigg[\max_{||\delta||_p \leq \epsilon} \mathcal{L}(x+\delta, y; \theta, \psi) \bigg],
\]
where $\epsilon$ is the perturbation budget that governs the adversarial robustness of the model. The model is trained to be robust in $\ell_p$-norm ball of radius $\epsilon$. Increasing $\epsilon$ means the model is stable for a larger radius. However, this framework is not appropriate for UDA as this requires labels and assumes data from a single domain. 

 \noindent Following \cite{madry2017towards}, we define the \textbf{adversarial robustness} as the accuracy of target dataset ($D_t$) perturbed with a perturbation budget of $\epsilon$ in  $\ell_{\infty}$-norm ball. To find the adversarial example $x_{adv}$, we use Projected Gradient Descent (PGD) with 20 iterations~\cite{madry2017towards}. \textit{We have used terms robustness and adversarial robustness interchangeably.}

\section{Pre-Training and Robustness}
\label{sec:pretraining}

\noindent We start with a simple question: can we instill robustness in unsupervised domain adaptation by replacing the normally pre-trained feature extractor with a robust counterpart?

\noindent To answer this question, we replaced the normal backbone model with an adversarially trained one. We call this setup \textbf{Robust Pre-Training} or Robust PT. To demonstrate the effect of robust pre-training, we conducted a set of experiments with six UDA methods and three common datasets, i.e., Office-31~\cite{saenko2010adapting}, Office-Home~\cite{venkateswara2017Deep} and VisDA-2017~\cite{peng2017visda}. We employed a ResNet-50~\cite{he2016deep} adversarially trained with different perturbation budgets as defined in Section~\ref{sec:preliminaries}. Unless stated otherwise, robustness is reported with PGD-20 and perturbation budget of $\epsilon=3$. For comparison, we use the default settings of all the hyper-parameters and report the average results over three independent runs. We only reported the best results averaged over all possible tasks of each dataset here. For detailed results, please refer to the supplementary material. 

\noindent It is reasonable to expect that adversarial pre-training will not increase robustness for unsupervised domain adaptation. Previous work has shown that the transferability of robustness is due to the robust feature representations learned by the pre-trained models. Robustness is only preserved if we do not update the backbone~\cite{hendrycks2019pretraining, shafahi2019adversarially}. Specifically, to maintain the robustness, only an affine layer is trained on top of the fixed feature extractor with the help of the labeled data. However, we fine-tuned the backbone model to be accurate in the source domain and invariant for the source and target domains.

\noindent  The best robustness results averaged over all tasks in each dataset are shown in Table~\ref{tab:pretraining_robustness}. We find that an adversarially pre-trained backbone can improve the robustness under UDA settings. For example, robustness for CDAN \cite{long2018conditional} improves from 0\% to 41.67\%, with around 5.5\% decrease in clean accuracy on VisDA-2017 dataset. For the DAN algorithm, improvement in robustness is 0\% to 22.11\% at the cost of an 18\% drop in clean accuracy. Similar improvement in robustness is also visible in experiments involving Office-31 and Office-Home datasets, as shown in Table~\ref{tab:pretraining_robustness}.

\noindent  However, adversarially pre-trained backbone decreases the generalization ability of models for the UDA setting. The decrease in accuracy can go as high as 20\%. We hypothesize that robust pre-training is not the most efficient way of leveraging robust features of the backbone. In the next section, we design an algorithm to utilize these features more efficiently. 

\begin{figure}
    \centering
    \includegraphics[width=1\linewidth]{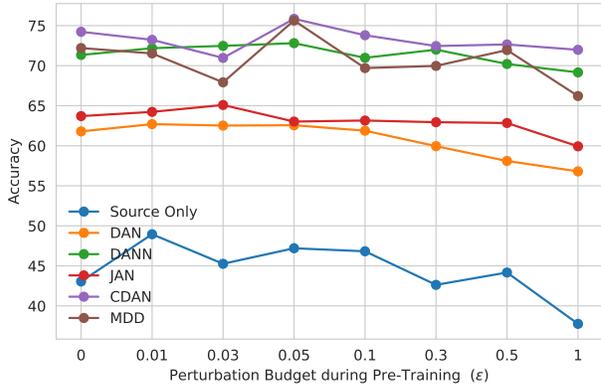}
    \caption{The clean accuracy of weak adversarially pre-trained (adversarial pre-training with small $\epsilon$) models on VisDA-2017 dataset. }
    \label{fig:student_model}
    \vspace{-10pt}
\end{figure}

\section{Robust Feature Adaptation}

\noindent In this section, we introduce our method and its motivation. The goal of Robust Feature Adaptation (RFA) is to improve the adversarial robustness of unsupervised domain adaptation (UDA) algorithms without causing a significant drop in accuracy. Based on our experiments in the previous section, we hypothesized that the direct use of pre-trained models as backbone model is not an efficient way to instill robustness in UDA training. These pre-trained models have significantly less accuracy to begin-with \cite{robustness}. This low pre-training accuracy makes it hard for UDA training to get better generalizations for the task. Our hypothesis is based on previous observations \cite{kornblith2019better} that have shown a direct relationship between the accuracy of a pre-trained model and its final performance on a given task.

\noindent In our method, we propose to adopt robust features instead of directly using robust models as a backbone. The main idea of the proposed method is to align the features of the UDA backbone model with the robust features provided by multiple adversarially pre-trained models. This aligning is done as we do domain adaptation training for learning domain invariant features.  

\noindent Each part of our framework is based on a hypothesis based on insights from previous works and detailed experimental investigation. In this section, we describe each component of our proposed algorithm along with their motivation. The empirical comparisons to support our method are given in Section~\ref{sec:design_principles}. An overview of the proposed method is illustrated in Figure~\ref{fig:block_rfa}.

\subsection{Feature Extractor for Domain Adaptation}
\noindent As described earlier, existing UDA algorithms fine-tune normally pre-trained ImageNet models. However, adversarially pre-trained models learn `fundamentally different features' compared to their normally pre-trained counterparts~\cite{tsipras2018robustness, engstrom2019adversarial, ilyas2019adversarial}. This difference can cause inconsistency between the features of student and teacher models, which may cause difficulty in optimization. Hence, we propose to use a weak adversarially pre-trained model (model pre-trained with a small perturbation budget) as the backbone model. 

\noindent As shown in Figure~\ref{fig:student_model}, these robust models do not hurt clean accuracy significantly but can solve the feature inconsistency problem. A experimental comparison is shown in Section~\ref{sec:design_principles}.

\begin{table}
\centering
\begin{adjustbox}{max width=0.5\textwidth, center}
\begin{tabular}{llcc}
    \toprule
    {Dataset}                    & {Method}    & {Accuracy}  & {Robustness}  \\
    \toprule
    \multirow{3}{*}{Office-31}   & Baseline   & 88.31     & 1.70  \\
                                 & Robust PT  & 80.72     & 67.54 \\
                                 & RFA        & 84.21     & 74.31 \\
    \hline
    \multirow{3}{*}{VisDA-2017}  & Baseline  & 72.20      & 4.03  \\
                                 & Robust PT & 67.72      & 39.50 \\
                                 & RFA       & 72.90      & 47.66 \\
    \hline
    \multirow{3}{*}{Office-Home} & Baseline  & 67.91      & 5.81  \\
                                 & Robust PT & 63.30      & 43.42 \\
                                 & RFA       & 65.37      & 51.13 \\

    \bottomrule
\end{tabular}
\end{adjustbox}
\caption{Comparison of robustness and clean accuracy for RFA with Robust Pre-Training and baseline. RFA improves robustness compare to Robust Pre-Training while keeping good generalization.}
\label{tab:main_results_datasets}
\end{table}

\begin{table}
\centering 
\begin{adjustbox}{max width=1\linewidth, center}
\begin{tabular}{llll}
    \toprule
     \multicolumn{1}{c}{UDA Method}      & Baseline    &  \multicolumn{1}{c}{Robust PT}     & \multicolumn{1}{c}{RFA} \\
     \midrule
     Source Only    & 43.05 / 0   & 25.67 / 6.64   & 44.65 / 11.10 \\
     DANN           & 71.34 / 0   & 65.79 / 38.21  & 65.32 / 34.11 \\
     DAN            & 61.79 / 0   & 42.24 / 22.11  & 55.70 / 21.59 \\
     CDAN           & 74.23 / 0   & 68.00 / 41.67  & 72.03 / 43.49 \\
     JAN            & 63.70 / 0   & 55.08 / 32.15  & 62.95 / 32.81 \\
    \bottomrule
\end{tabular}
\end{adjustbox}
\caption{Comparison of Robust Pre-Training and RFA for five UDA algorithms with the VisDA-2017 dataset. RFA significantly improves robustness while keeping good clean accuracy.}
\label{tab:rfa_visda_robustness}
\end{table}


\begin{table*}
\centering
\begin{tabular}{ccc}
    
    \begin{adjustbox}{width=0.2\linewidth}
    
    \begin{tabular}{lll}
        \toprule
         Student       & Acc.    &  Rob.   \\
         \toprule
         Baseline      & 72.20 & 4.03    \\
         Normal        & 71.22 & 7.63    \\
         Adv.          & 72.71 & 40.61  \\ 
        \bottomrule
    \end{tabular}
    
    \end{adjustbox}
    &
     \begin{adjustbox}{width=0.4\linewidth}
    \begin{tabular}{lccccc}
        \toprule
     Loss       & DANN               & CDAN                       & MDD  \\
        \toprule
        L1                 & 45.02 / 9.58       & 55.16 / 13.53              & 54.52 / 18.89         \\
        L2                 & 54.28 / 1.45      & 58.16 / 1.76                & 64.20 / 8.29          \\
        SP                 & 65.32 / 34.11      & 72.03 / 43.49              & 72.90 / 47.66          \\
        \bottomrule
    \end{tabular}
    \end{adjustbox}
    
    &
    \begin{adjustbox}{width=0.33\linewidth}
    \begin{tabular}{lll}
        \toprule
        Method          & \multicolumn{1}{c}{RN-18}             & \multicolumn{1}{c}{WRN-50-2}      \\
        \toprule
        Baseline        & 69.61 / 0.15      & 73.36 / 5.47  \\
        Robust PT       & 64.44 / 24.40     & 71.20 / 37.63 \\
        Ours (RFA)      & 65.05 / 36.46     & 74.98 / 50.47 \\
        \bottomrule 
    \end{tabular}
    \end{adjustbox} \\
    (a)&(b)&(c)\\
\end{tabular}
\caption{(a) Comparison of normally and adversarially pre-trained students for the accuracy and robustness of our algorithm. (b) Comparison of pairwise loss vs. similarity preserving loss for robustness. (c) Comparison of accuracy and robustness (\%) for MDD Baseline, Robust PT and RFA with different neural network architectures. RFA consistently improves robustness for different architectures. Here RN represents ResNet and WRN WideResNet. All of these experiments are conducted on VisDA-2017 dataset.}
\label{tab:ablation_studies_set2}
\end{table*}

\subsection{Joint Training for Adaptation of Robust and Domain Invariant Features}

\noindent Our robust feature adaptation method aims to fine-tune the UDA feature extractor in such a way that it adapts robust features from adversarially trained models along with domain-invariant features from UDA training. 

\noindent In knowledge distillation, we initialize the student with random weights and force the student to mimic the feature space of the teacher by minimizing the pair-wise distance between features and/or softened class scores.
Our UDA feature extractor, on the other hand, is also pre-trained and has already learned a set of features. This means that the student and the teacher may have learned features in different ways, or the order of the learned feature maps may differ. 
Furthermore, \textit{since the teacher is not trained directly on the target dataset, it can not provide the softened class scores.} This is also another reason not to directly minimize pair-wise distance as the teacher is trained on a different dataset. In conclusion, we only want to use the feature supervision of the teacher to align student's features with it to adapt robustness.

\noindent To align features of student to that of robust teacher, we used similarity preserving loss to match the similarity of activations between robust and non-robust features ~\cite{tung2019similarity}. The main idea of this loss is to align the student's feature in such a way that two inputs producing similar activations in the feature space of teacher model should also produce similar activations in the feature space of student model. Specifically, given a mini-batch of training data, let $Q_{T}^{l} \in \mathbb{R}^{b \times d}$ and $Q_{S}^{l} \in \mathbb{R}^{b \times d}$ denote the activations of $l$-th layer from teacher and student models, respectively, where $b$ is the batch size and $d$ is the dimension of the activations after reshaping. The similarity matrices of $l$-th layer from teacher and student models are defined as $G_{T}^{l} = Q_{T}^{l} \cdot {Q_{T}^{l}}^\intercal / ||Q_{T}^{l} \cdot {Q_{T}^{l}}^\intercal||_2$ and $G_{S}^{l} = Q_{S}^{l} \cdot {Q_{S}^{l}}^\intercal / ||Q_{S}^{l} \cdot {Q_{S}^{l}}^\intercal||_2$, respectively, where $||\cdot||_2$ is a row-wise $\ell_2$-norm. We then define the robust feature adaptation loss of $l$-th layer as
\[
\mathcal{L}_{RFA}^l = \dfrac{1}{b^2} ||G_{T}^{l} - G_{S}^{l}||_F^2,
\label{eq:1}
\]
where $||\cdot||_F$ is the Frobenius norm.

\noindent We use the sum of robust feature adaptation losses of intermediate layers:
\begin{equation}\label{equ:rfa}
\mathcal{L}_{RFA} = \sum_{l=1}^{L} \mathcal{L}_{RFA}^l,
\end{equation}
where $L$ is the number of intermediate layers. The joint training loss is then defined as
\begin{equation}\label{equ:joint_loss}
\mathcal{L} = \mathcal{L}_{C} + \mathcal{L}_{DA} + \alpha \mathcal{L}_{RFA},
\end{equation}

where $\mathcal{L}_{C}$ is the classification loss on the source domain, $\mathcal{L}_{DA}$ is the loss term for domain adaptation, and $\alpha$ is a hyper-parameter that balances domain adaptation and robust feature adaptation. Note that our proposed method can be applied to different UDA algorithms by using the corresponding domain adaptation method with loss term $\mathcal{L}_{DA}$.

\subsection{Adapting Diverse Robust Features}
\begin{figure}
    \centering
    \includegraphics[width=0.45\textwidth]{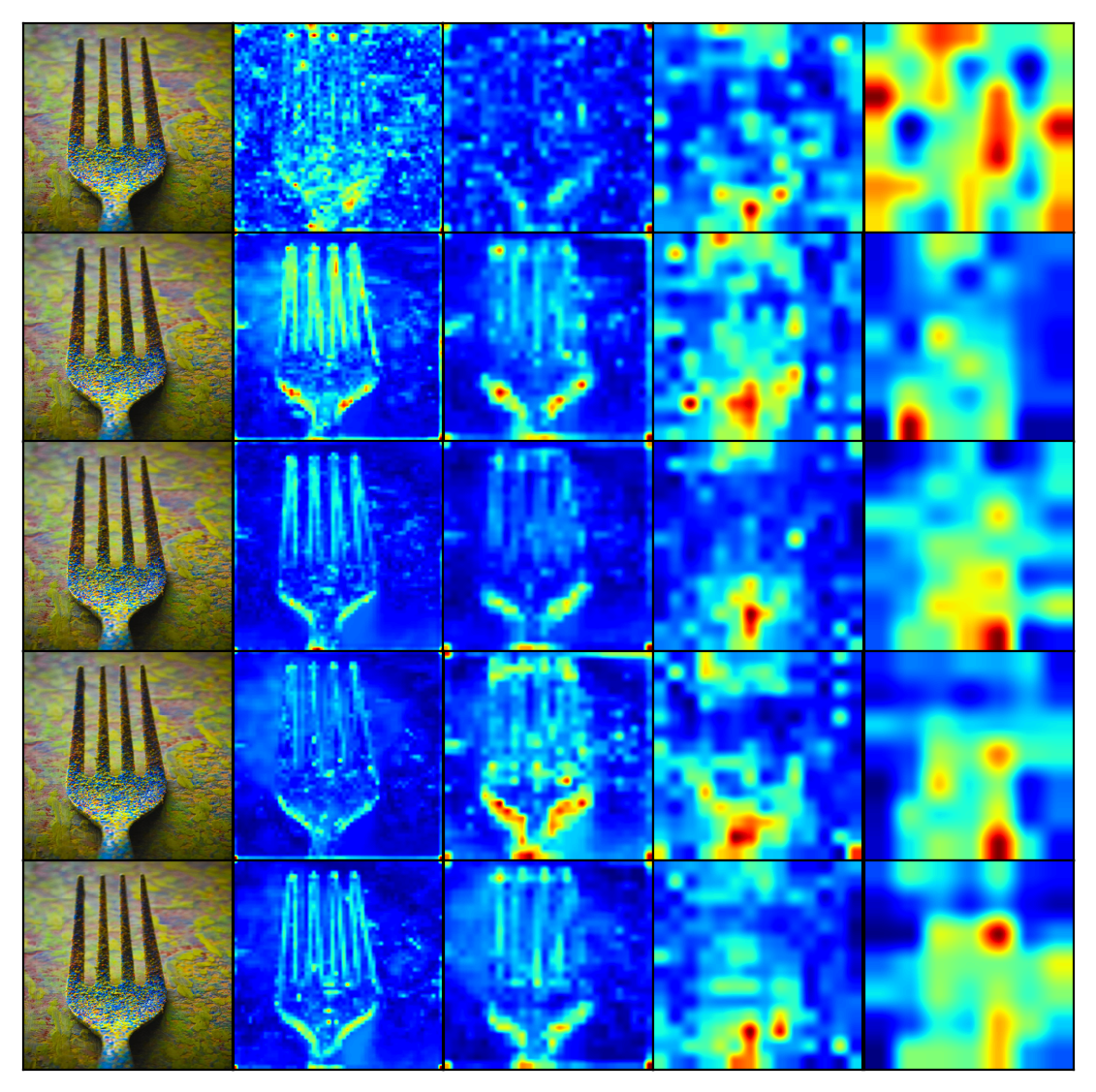}
    \caption{Maximally activated neurons for an image from Office-Home dataset. \textit{The first row shows activations for normally pre-trained model} and other rows show activations for robust pre-trained models trained with a different perturbation budget ($\epsilon$). Highlighted regions can be interpreted as the discriminative parts of the input that activates the neurons the most. Note that different models have learned different discriminative features.}
    \label{fig:activatons}
\end{figure}
\noindent The Figure~\ref{fig:activatons} shows the diversity of discriminative features learned by the same model trained with different perturbation budgets. More details are in Section~\ref{sec:design_principles}. To leverage these diverse robust features, we propose to supervise the student with multiple teachers. To reduce the computing cost during training, we randomly choose one teacher at each iteration during training. This means that we can guide the student model with the diversity of multiple teachers with the same computing cost as using one. The detailed procedure for our method is summarized in Algorithm \ref{algo1}.

\begin{algorithm}
	\caption{RFA: Robust Feature Adaptation for UDA}
	\label{algo1}
	\begin{algorithmic}[1]
		\REQUIRE Multiple robust teachers $\{f(\cdot\ ;\theta^m_T)\}_{m=1}^{M}$, training datasets $D_s, D_t$, batch size $b$, learning rate $\eta$, hyperparameter $\alpha$, iteration number $K$, UDA algorithm.
		\ENSURE $\theta_S$, $\psi_S$.
		\STATE Initialize $\theta_S$, $\psi_S$;
		\FOR {$0 \leq k \leq K-1$} 
		\STATE Sample a random mini-batch of training examples $\{((x^s_i, y^s_i), (x^t_i))\}_{i=1}^b$ with a batch size of $b$;
		\STATE $x \gets \{(x^s_i, x^t_i)\}_{i=1}^b$;
		\STATE Randomly sample a teacher $f(\cdot\ ;\theta^m_T)$;
		\STATE Compute $Q^l_T$ and $Q^l_S$ with $f(x;\theta^m_T)$ and $f(x;\theta_S)$ respectively, for $l=1,2,\cdots, L$; 
		\STATE Compute $\mathcal{L}_{RFA}$ according to Eq.~\eqref{equ:rfa};
		\STATE Compute $\mathcal{L}_{C}$ and $\mathcal{L}_{DA}$ with the UDA algorithm;
		\STATE Compute $\mathcal{L}$ according to Eq.~\eqref{equ:joint_loss};
		\STATE Update $(\theta_S, \psi_S) \gets (\theta_S, \psi_S) - \eta \cdot \nabla_{(\theta_S, \psi_S)}\mathcal{L}$;
		
		\ENDFOR
	\end{algorithmic}
\end{algorithm}

\section{Experiments}

\subsection{Setup}
\noindent We conduct experiments on 19 different tasks derived from 3 main-stream unsupervised domain adaption (UDA) datasets.
\textbf{Office-31}~\cite{saenko2010adapting} is a standard domain adaptation dataset with 6 tasks based on three domains: Amazon (\textbf{A}), Webcam (\textbf{W}) and DSLR (\textbf{D}). The dataset is imbalanced across domains with 2,817 images in \textbf{A}, 795 images in \textbf{W} and 498 images in \textbf{D} domain. \textbf{Office-Home}~\cite{venkateswara2017Deep} is a more complex dataset compared to Office-31 and contains more images (15,500) for 12 adaptation tasks based on 4 more diverse domains: Artistic (\textbf{Ar}), Clip Art (\textbf{Cl}), Product (\textbf{Pr}), and Real World (\textbf{Rw}). \textbf{VisDA-2017}~\cite{peng2017visda} is a simulation-to-real dataset with two extremely different domains: synthetic domain in which images are collected from 3D rendering models and real-world images. It is also a large-scale dataset as it contains 280k images in the synthetic domain and 50k images in the real-world domain. Due to the extremely different domains and scale, it is one of the most challenging datasets in UDA.

\noindent Unless stated otherwise, we use ResNet-50~\cite{he2016deep} as our backbone model and MDD~\cite{zhang2019bridging} as the domain adaptation algorithm. We used this setup to show that our method can improve robustness without a significant drop in accuracy. To show that Robust Feature Adaptation (RFA) can work as a plug-in method, we conduct experiments with six UDA algorithms: Source Only (fine-tuning model on source data only), DAN~\cite{long2015learning}, DANN~\cite{ganin2016domain}, JAN~\cite{long2017deep}, CDAN~\cite{long2018conditional}, and MDD~\cite{zhang2019bridging}. We follow the experimental protocol of~\cite{ganin2016domain, long2018conditional} commonly used in UDA and adopt the hyper-parameters used in~\cite{dalib}. We compare RFA with \textbf{UDA algorithm Baseline} (adopting normally pre-trained ImageNet model) and \textbf{Robust PT} (UDA algorithm adopting adversarially pre-trained ImageNet model). For a fair comparison, we use the same values for all hyper-parameters for the UDA algorithm Baseline, Robust PT, and RFA. The new hyper-parameter of our proposed method is $\alpha$. We choose it based on the magnitude of domain adaptation loss. Specifically, we multiply robust feature adaptation loss $\mathcal{L}_{RFA}$ by $1000$ to make it have the equivalent magnitude to that of domain adaptation loss. We report average results over three runs for all the experiments. 

\subsection{Results}
\noindent\textbf{On Improving Robustness.} To achieve better robustness, we choose four strong teachers, i.e., ImageNet ResNet-50 models, trained with different perturbation budgets. More specifically, we use perturbation budget of $\epsilon \in \{3, 5\}$ with $\ell_2$-norm and $\epsilon \in \{2, 4\}$ with $\ell_{\infty}$-norm. To show the effectiveness of our method, we choose a backbone adversarially trained with $\epsilon=1$. For the bulk of our experiments, We use MDD as a domain adaptation algorithm.

\noindent The average results for Office-31, Office-Home, and VisDa-2017 are shown in Table~\ref{tab:main_results_datasets}. These results clearly show that our method can improve the robustness of the backbone model by adapting the robust features without a significant drop in clean accuracy. The improvement in robustness is due to the robust teachers, while the improvement in clean accuracy is because of the backbone model used in RFA. This model has higher accuracy compared to backbone use in Robust Pre-Training. This way, our method has a significant advantage over Robust PT as it can use backbone models with higher clean accuracy while adapting robustness from any teacher.

\begin{table*}
\begin{adjustbox}{ width=2.1\columnwidth, center}
\centering
\begin{tabular}{lccccccccccccc}
    \toprule
    Method                      & Ar $\shortrightarrow$ Cl & Ar $\shortrightarrow$ Pr & Ar $\shortrightarrow$ Rw & Cl $\shortrightarrow$ Ar & Cl $\shortrightarrow$ Pr & Cl $\shortrightarrow$ Rw & Pr $\shortrightarrow$ Ar & Pr $\shortrightarrow$ Cl & Pr $\shortrightarrow$ Rw & Rw $\shortrightarrow$ Ar & Rw $\shortrightarrow$ Cl & Rw $\shortrightarrow$ Pr & Avg \\
    \toprule
    Baseline   & 54.59    & 72.38    & 77.19   & 61.52       & 71.19      & 71.54     & 63.04    & 50.31    & 79.0   & 72.5   & 57.66   & 83.92     & 67.91  \\
    Robust PT    & 55.07    & 73.87    & 78.26   & 60.82       & 71.84      & 71.88     & 60.65    & 51.89    & 79.02  & \textbf{72.64}  & \textbf{60.50}    & 82.81    & 68.27   \\
    Ours (RFA)     & \textbf{55.65}    & \textbf{77.13}    & \textbf{80.69}   & \textbf{64.43}       & \textbf{74.81}      & \textbf{75.54}     & \textbf{63.99}    & \textbf{53.07}    & \textbf{80.59}  & 71.80  & 58.41    & \textbf{84.31}    & \textbf{70.03}   \\
    \bottomrule
\end{tabular}
\end{adjustbox}
\caption{Classification accuracy (\%) for all the twelve tasks from Office-Home dataset based on ResNet-50. Our method improves clean accuracy of 10 out of 12 tasks as well as the average.}
\label{tab:rfa_officehome_clean}
\vspace{-10pt}
\end{table*}

\begin{table}
\centering
\begin{tabular}{cc}
    \begin{adjustbox}{width=0.55\linewidth}
    \begin{tabular}{llccccc}
    \toprule
    $\alpha$ & 100 & 500 & 1000& 5000            \\
    \toprule
    Acc.& 71.61 & 73.62 & 72.90 & 70.31  \\ 
    Rob. & 40.07 & 46.36 & 47.66 & 47.27  \\
    \bottomrule
    \end{tabular}
    \end{adjustbox}
    &
    \begin{adjustbox}{width=0.4\linewidth}
    \begin{tabular}{lcc}
    \toprule
     Teachers & Acc.  & Rob.             \\
    \toprule 
    Single  & 70.31 & 40.15 \\ 
    Multiple & \textbf{73.45} & \textbf{40.87} \\ 
    \bottomrule
    \end{tabular}
    \end{adjustbox}\\
    (a) & (b)\\
\end{tabular} 
\caption{\textbf{Ablation Studies.} (a) The effect of varying $\alpha$ on accuracy and robustness (\%) for RFA on VisDA-2017 dataset. (b) The effect of multiple teachers on accuracy and robustness (\%) on VisDA-2017 dataset.}
\label{tab:ablation_studies_set1}
\end{table}

\noindent \textbf{On RFA as a Plug-in Method.} A salient characteristic of our method is that it can complement existing or new domain adaption algorithms that use ImageNet pre-trained models. To show this, we conduct experiments with six different UDA algorithms (Source only, DAN, DANN, JAN, CDAN, and MDD) on the challenging and large-scale VisDA-2017 dataset. As shown in Table~\ref{tab:rfa_visda_robustness}, RFA improves robustness for all the six UDA algorithms.

\begin{figure}
    \centering
    \includegraphics[ width=0.5\textwidth]{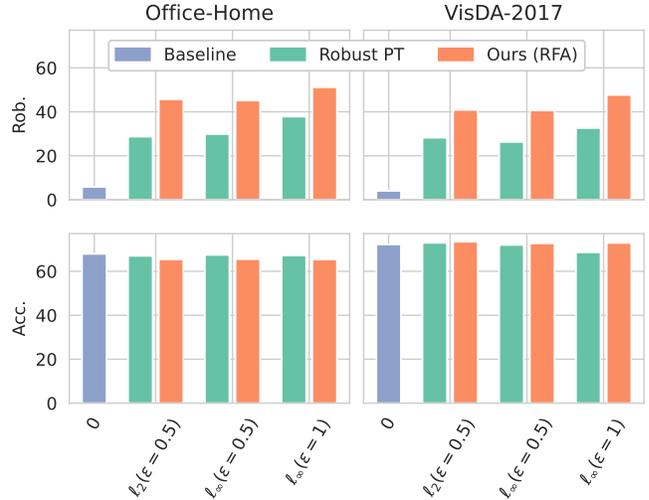}
    \caption{Comparison of MDD Baseline, Robust PT (Pre-Training), and RFA for average robustness and accuracy (\%) on Office-Home and VisDA-2017. The $x$-axis shows the perturbation budget of the pre-trained model.}
    \label{fig:rfa_robustness_avg}
\end{figure}

\section{Discussion and Analysis}
\subsection{Empirical Investigation of Design Principles}
\label{sec:design_principles}
\noindent \textbf{Choosing Student Model.}
One major insight of our framework is the use of weak adversarially pre-trained models (adversarially pre-trained models with small perturbation budget $\epsilon$) as feature extractors. To see the effect of the weak adversarially pre-trained model, we compare it with a normally pre-trained student in Table~\ref{tab:ablation_studies_set2}(a). Normally pre-trained student can improve robustness, albeit not significantly. Weak adversarially pre-trained students, on the other hand, can improve robustness significantly.

\noindent To further see how the UDA feature extractor model should be pre-trained, we compare the robustness and accuracy of different feature extractor models with different pre-training perturbation levels in Figure~\ref{fig:rfa_robustness_avg}. 

\noindent \textbf{Comparison of Pairwise and with Non-Pairwise Loss.}
An important aspect of our algorithm is the loss function. We hypothesized that similarity preserving loss that preserves similarity between the activations is better to compare to pair-wise loss. This is because our student model is already trained, and we only want to fine-tune it and require weak supervision. To illustrate it, we compare the robustness and clean accuracy for two pair-wise losses with similarity preserving loss in Table~\ref{tab:ablation_studies_set2}(b).

\noindent\textbf{Effect of Multiple Teachers.} We hypothesized that the same model trained with different perturbation budgets can supervise student models with the diverse features. In Figure~\ref{fig:activatons}, we show the maximally activated neurons (maximum value across channels) of four different residual blocks of the robust ResNet-50 model. The first row shows activations of residual blocks for a normally pre-trained model, and other rows represent activations for robust ResNet-50 models trained with different values of $\epsilon$. The figure shows the diversity of discriminative features learned.

\noindent To illustrate the effect of multiple teachers, we compare it with a single teacher in Table~\ref{tab:ablation_studies_set1}(b). Single model supervision is enough to distill the robustness. However, the diversity of supervision from multiple teachers improves both accuracy and robustness.

\subsection{Ablation Studies}
\noindent\textbf{Sensitivity of Weight of Robust Feature Adaptation ($\alpha$).} We study the sensitivity of our method to the weight of robust feature adaptation term $\alpha$ on VisDA-2017. Table~\ref{tab:ablation_studies_set1}(a) demonstrates the clean accuracy and adversarial robustness by varying $\alpha \in \{0, 100, 500, 1000, 5000\}$. Increasing $\alpha$ decreases the clean accuracy while increasing the robustness. These experimental results show that $\alpha$ can control the trade-off between clean accuracy and adversarial robustness.


\begin{table}
\centering
\begin{adjustbox}{width=0.5\textwidth, center}
\begin{tabular}{lcccccc}
    \toprule
    \multirow{2}{*}{Method}             &\multirow{2}{*}{Clean}             & \multirow{2}{*}{FGSM}        & \multicolumn{4}{c}{PGD-k}\\
    \cline{4-7}
               &         &                  & 10& 20 & 50 & 100    \\
    \toprule
    Baseline       & 72.20 & 41.15  & 11.82                & 4.03  & 3.24 & 3.06   \\ 
    Robust PT    & 71.95 & 63.23  & 39.54               & 28.21 & 25.55 & 24.69   \\ 
    Ours     & \textbf{73.45} & \textbf{67.87} & \textbf{42.25} & \textbf{40.87} & \textbf{40.28} & \textbf{40.11}   \\ 
    \bottomrule
\end{tabular}
\end{adjustbox}
\caption{The effect of an increasing number of iterations for PGD attack. Results of the proposed method are consistent, showing a successful convergence of PGD attacks.}
\label{tab:iterations}
\end{table}


\noindent\textbf{Effect of the number of PGD iterations on robustness.} To further show the transferability of robustness, we test our method with an increasing number of iterations for PGD attack (PGD-k). The robustness of our method is consistent as shown in Table~\ref{tab:iterations}. 

\begin{table*}
\centering
\begin{tabular}{cc}
    \begin{adjustbox}{width=0.51\linewidth}
    \begin{tabular}{lccccccc}
        \toprule
        Method      & A $\shortrightarrow$ W   & D $\shortrightarrow$ W   & W $\shortrightarrow$ D   & A $\shortrightarrow$ D   & D $\shortrightarrow$ A   & W $\shortrightarrow$ A   & Avg. \\
        \toprule
         Baseline         & 91.40      & 98.74    & 100.00     & 92.17      & 73.06      & 74.47     & 88.31 \\
        Robust PT    & 91.78      & 99.12    & 100.00     & 92.77      & 73.85      & 74.11     & 88.60\\
        Ours (RFA)   & \textbf{92.80}      & \textbf{99.21}    & \textbf{100.00}     & \textbf{93.04}      & \textbf{78.00}      & \textbf{77.74}     & \textbf{90.15} \\
        \bottomrule
    \end{tabular}
    \end{adjustbox}
    &
    \begin{adjustbox}{width=0.46\linewidth}
    \begin{tabular}{lcccccc}
        \toprule
        Method                                   & Source     & DANN             & DAN              & CDAN              & JAN               & MDD  \\
        \toprule
        Baseline                                      & 43.05         & 71.34       & 61.79      & 74.23         & 63.70         & 72.20    \\
        Robust PT               & 47.20         & 72.81       & 62.56      & 75.85         & 63.02         & 75.64  \\
        Ours (RFA)                  & \textbf{59.00}         & \textbf{75.05}       & \textbf{65.58}      & \textbf{77.54}         & \textbf{66.68}         & \textbf{79.42}\\
        \bottomrule
    \end{tabular}
    \end{adjustbox}
    \\
    (a)&(b)\\
     
\end{tabular}
\caption{\textbf{Improved Clean Accuracy.} (a) Classification accuracy (\%) for all the six tasks from Office-31 dataset based on ResNet-50. (b) Comparison of classification accuracy (\%) for Baseline, Robust PT and RFA with six UDA algorithms on VisDA-2017 dataset. RFA consistently improves accuracy for all UDA algorithms.}
\label{tab:office31-clean_acc}
\end{table*}

\begin{table}
\centering 
\begin{adjustbox}{width=1\linewidth, center}
    \begin{tabular}{lcccc|c}
        \toprule
        Method    & Art-Painting  & Cartoon      & Sketch         & Photo        & Average   \\
        \hline
        \multirow{2}{*}{Baseline}    & 77.93         & 80.29        & 78.90          & 94.55        & 82.92    \\
                  & \rob{0}       & \rob{0.13}   & \rob{2.24}     & \rob{0.18}   & \rob{0.64} \\
        \hline
        \multirow{2}{*}{Ours (RFA)}        & 76.56         & 76.83        & 75.97          & 94.61        & 81.00     \\
                  & \rob{23.15}   & \rob{51.58}  &  \rob{62.82}   & \rob{40.00}  & \rob{44.38} \\
        \bottomrule
    \end{tabular}
\end{adjustbox}
\caption{Comparison of accuracy and \rob{robustness} (\%) for DecAug Baseline, Robust PT and RFA for all the four tasks from PACS based on ResNet-18.}
\label{tab:rfa_decaug_pacs_robustness}
\end{table}

\begin{table}
\centering
\begin{adjustbox}{width=0.5\textwidth, center}
\begin{threeparttable}
\begin{tabular}{lcccccccc}
    \hline
    Dataset    & Rob. & Source  & DANN             & DAN             & CDAN           & JAN            & MDD           \\
               & PT  & Only   & \cite{ganin2016domain} & \cite{long2015learning}             & \cite{long2018conditional}          & \cite{long2017deep}            & \cite{zhang2019bridging}           \\
    \hline
    VisDA &$\times$     & 43.05   & 71.34       & 61.79     & 74.23     & 63.70     & 72.20    \\
    2017           &$\checkmark$       & \textbf{48.95}     & \textbf{72.81}      & \textbf{62.70}    & \textbf{75.85}     & \textbf{65.51}    & \textbf{75.64}  \\
    \hline
     Office& $\times$               & \textbf{77.80}               & 85.79               & 81.72               & 86.90            & 85.68                       & 88.31    \\
        31      & $\checkmark$           &  77.66                       & \textbf{86.06}      & \textbf{82.08}      & \textbf{88.05}   & \textbf{86.05}              & \textbf{88.60}  \\

    \hline
    Office & $\times$        & 58.29      & 63.39                & 59.64                       & 67.03             & 64.61                       & 67.91    \\
     Home             & $\checkmark$    & \textbf{58.87}     & \textbf{64.08}        & \textbf{60.38}               & \textbf{67.67}   & \textbf{65.60}              & \textbf{68.27}  \\

    \hline
\end{tabular}
\begin{tablenotes}
	\item $\times$: Normally Pre-Trained Model, $\checkmark$: Adversarially Pre-Trained Model, Rob. PT: Robust Pre-Training.
\end{tablenotes}
\end{threeparttable}
\end{adjustbox}
\caption{Comparison between normally and adversarially pre-trained models on classification accuracy (\%) with different UDA algorithms. Adversarial pre-training improves classification accuracy for UDA.
}
\label{tab:pretraining_clean}
\vspace{-15pt}
\end{table}

\noindent\textbf{Improvement by RFA is consistent across architectures.}
In Table~\ref{tab:ablation_studies_set2}(c), we demonstrate that our proposed method can improve robustness using different architectures. RFA improves the robustness of Wide-ResNet-50-2 from 5.47\% to 50.47\% and accuracy of ResNet18 from 0.15\% to 36.46\%. 

\subsection{Can RFA Improve Robustness for Domain Generalization?}
\noindent An important aspect of our method is that it is domain-agnostic and can be applied to tasks involving more than one domain. To illustrate this, we also conduct experiments for Domain Generalization (DG) with our method on PACS~\cite{Li2017} dataset. DG methods~\cite{Li2017,carlucci2019domain,zhou2020learning,bai2020decaug} learn models from multiple domains such that they can generalize well to unseen domains. PACS dataset contains four domains with different image styles: art painting, cartoon, sketch, and photo. We follow the same leave-one-domain-out validation experimental protocol as in~\cite{Li2017}. For each time, we select three domains for training and the remaining domain for testing. We apply RFA to the SOTA DG method DecAug~\cite{bai2020decaug} and report results in Table~\ref{tab:rfa_decaug_pacs_robustness}. It illustrates that our method can also significantly improve the robustness while maintaining good clean accuracy in domain generalization.

\subsection{Can Adversarially Pre-Trained Models Improve Clean Accuracy?}
\noindent A recent work~\cite{salman2020adversarially} has shown that weak adversarially pre-trained models (AT with small $\epsilon \in [0.01,0.5]$) can also improve clean accuracy for target tasks in transfer learning, e.g., ImageNet to Pets dataset. In this section, we explore this hypothesis for unsupervised domain adaptation (UDA). Specifically, we did experiments for two settings: using weak adversarially pre-trained models as feature extractors and using them as teachers in our proposed algorithm.

\noindent First, we use a weak adversarially pre-trained model as a feature extractor while keeping everything else the same as in UDA training. We found that this simple setup can improve clean accuracy. The results are shown in Table~\ref{tab:pretraining_clean}. 

\noindent To further see the effect of robust features, we used these weak adversarially trained models in our robust adaptation algorithm. The results on different tasks from Office-31, Office-Home and average accuracy for different UDA algorithms on VisDA-17 are shown in Tables \ref{tab:office31-clean_acc}(a),\ref{tab:rfa_officehome_clean}, \ref{tab:office31-clean_acc}(b), respectively. RFA outperforms both Baseline and Robust Pre-Training with significant margins. Our method achieves 90.15\% compared to 88.31\% of Baseline and 88.60\% of Robust Pre-Training on Office-31. Similarly, on a more complex Office-Home dataset, it achieved 70.03\% compared to 67.91\% of Baseline and 68.27\% of Robust PT. On challenging the VisDA-2017 dataset, we achieved even more improvements. For instance, MDD with normally pre-trained ResNet-50 achieves an accuracy of 72.20\%, but our proposed algorithm achieves 79.42\% -- an absolute 7\% improvement. 

\noindent It is noteworthy that our method significantly \textit{improves accuracy on hard tasks}, e.g., for Office-31, \textbf{D} $\to$ \textbf{A} (73.06\% to 78\% ) and \textbf{W} $\to$ \textbf{A} (74.47\% to 77.74\% ); for Office-Home, \textbf{Cl} $\to$ \textbf{Ar} (61.52\% to 64.43\%),  \textbf{Cl} $\to$ \textbf{Pr} (71.19\% to 74.81\%) and \textbf{Cl} $\to$ \textbf{Rw} (71.54\% to 75.54\%); for VisDA-2017, simulation to real (72.20\% to 79.42\%). This highlights the importance of adaptation of robust features for UDA.

\section{Conclusion}
\noindent Existing interventions for adversarial robustness require labels and assume learning from a single domain. This hinders their application in unsupervised domain adaptation. To make unsupervised domain adaptation robust, we introduced a simple, unsupervised and domain-agnostic method that does not require adversarial examples during training. Our method is motivated by the transferability of robustness. It utilizes adversarially pre-trained models and adapts robustness from their internal representations. Our results show that it significantly improves the robustness for UDA. 

\noindent Our work may be extended in two dimensions. One direction is the applicability of our work in other problems involving learning from multiple domains. In this work, we primarily focused on UDA and also briefly discussed domain generalization. However, many different scenarios require learning from a diverse set of domains, such as open compound domain adaptation~\cite{liu2020open} and single domain generalization~\cite{qiao2020learning}. It would be interesting to see the performance of our algorithm under these circumstances. Another direction is to leverage a zoo of diverse pre-trained models~\cite{xu2021nasoa, shu2021zoo}. Systematic selection of relevant teachers and adaptive aggregation of their knowledge can further improve performance without significant computation overhead.

\noindent \textbf{Acknowledgements.} Authors are thankful to the anonymous reviewers, and to Dr. Nauman, Faaiz, Teerath, Salman and Asim for their help and constructive feedback. 
\FloatBarrier

{\small
\bibliographystyle{ieee_fullname}
\bibliography{egbib}
}

\begin{appendices}
\section{Extended Experiemnt Details}

\noindent \textbf{Experimental Setup}: Unless stated otherwise, we use ResNet-50 as the backbone model. We follow~\cite{dalib}'s experimental design. Specifically, we use 1) learning rate of 0.01; 2) a batch size of 64: 32 for source and 32 for target domain; 3) 30 epochs training; 4) 1000 iterations of data per epoch; 5) data augmentation: random horizontal flip for all datasets and center crop for VisDA-2017. We run each experiment 3 times and report the average value of both clean accuracy and robustness. All the experiments are implemented in PyTorch.

\noindent We also use the same configuration for our proposed method (RFA). The only difference is that we forward passed input batch of data through one frozen teacher model per iteration and get intermediate activation to adapt robust features. We use three datasets for our UDA experiments: VisDA-2017, Office-31, and Office-Home. 

\noindent \textbf{Extended Results for Robust Pre-Training}: We provide extended results for Section 4 of our main paper. Specifically, we replaced the normally pre-trained ResNet-50 model with adversarially robust ResNet-50 models pre-trained with different perturbation budgets ($\epsilon$) on ImageNet. We conduct experiments with six UDA algorithms on aforementioned datasets. The results averaged over all possible tasks of each dataset are reported in Table~\ref{tab:pretraining_extended}. The robustness is tested with a PGD-20 attack and perturbation budget of $\epsilon=3$. The results show that merely replacing the pre-trained model with the robust model can improve the robustness, but it also causes a significant drop in clean accuracy.

\begin{figure*}
    \centering
    \includegraphics[ width=1.0\textwidth]{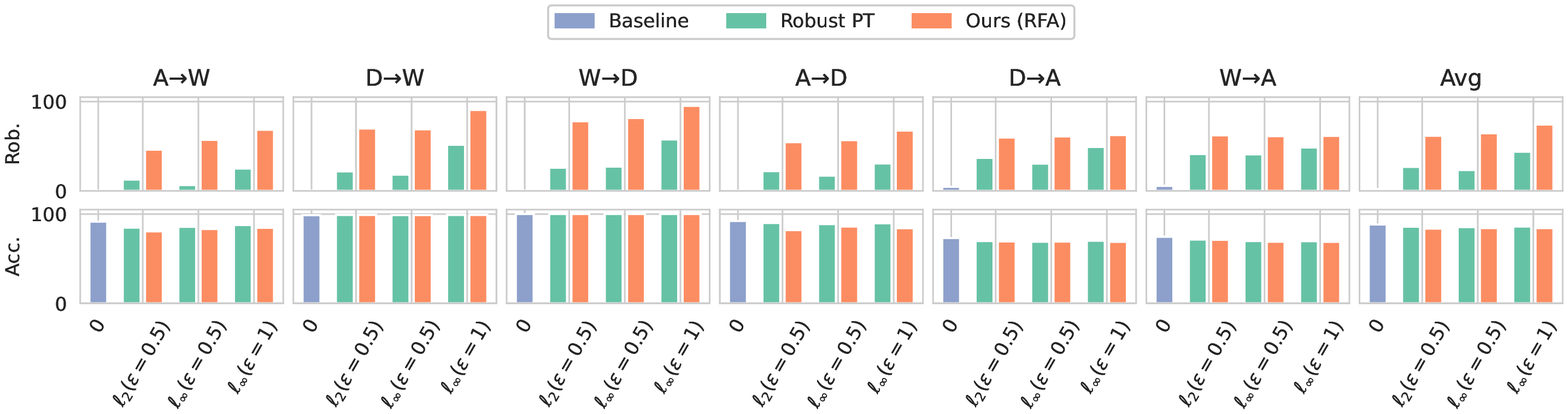}
    \caption{Comparison of robustness and accuracy (\%) for MDD Baseline, Robust PT and RFA on six tasks from Office-31 dataset. The $x$-axis is the perturbation budget of the pre-trained model. RFA consistently improves robustness with a small drop in the clean accuracy.}
    \label{fig:rfa_robustness_office31}
\end{figure*}

\begin{figure*}
    \centering
    \includegraphics[width=1\textwidth]{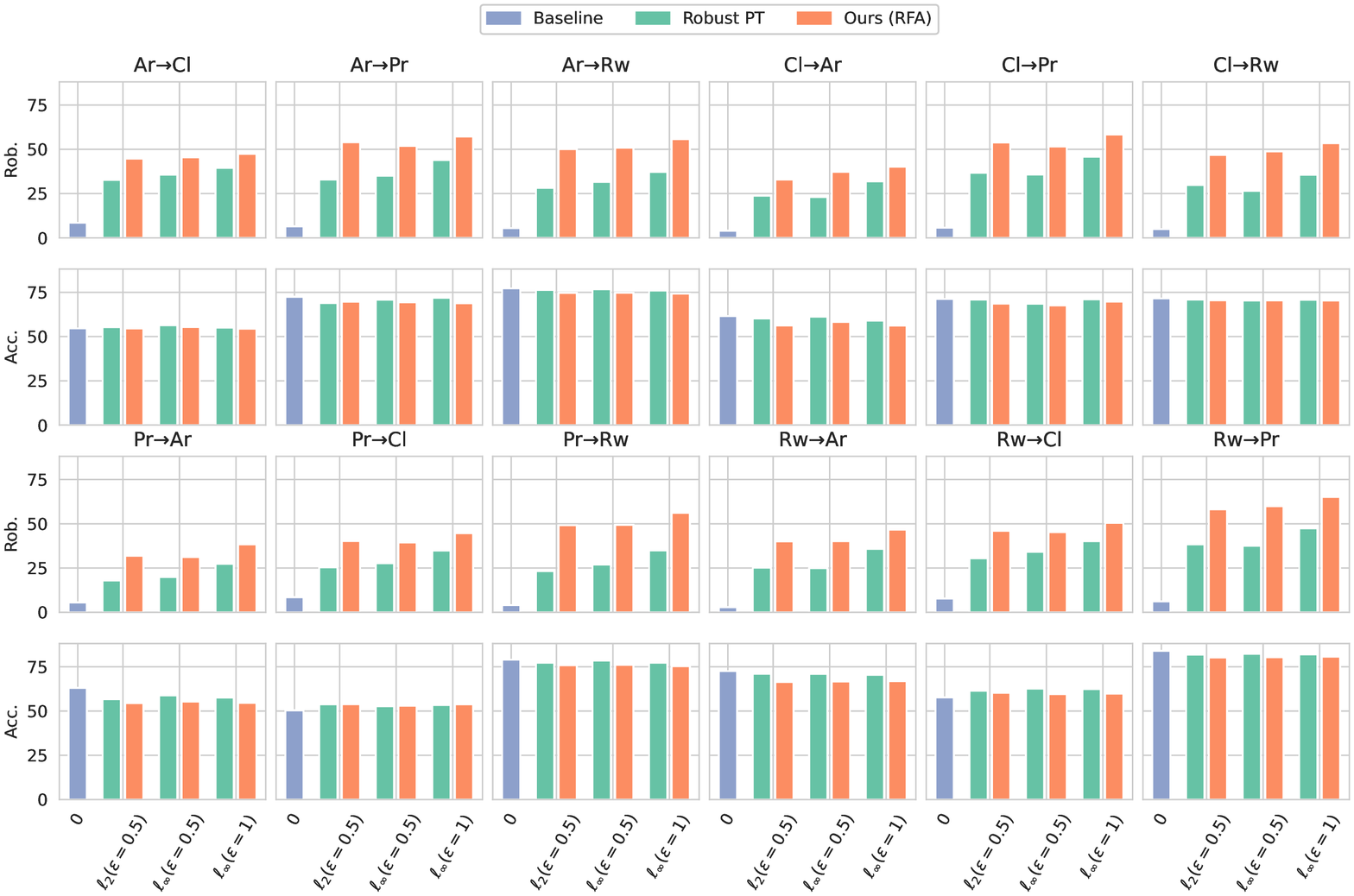}
    \caption{Comparison of robustness and accuracy (\%) for MDD Baseline, Robust PT and RFA on twelve tasks from Office-Home dataset. The $x$-axis is the perturbation budget of the pre-trained model. RFA consistently improves robustness with a small drop in the clean accuracy.}
    \label{fig:rfa_robustness_office_home}
\end{figure*}

\noindent \textbf{Extended Results for Our Method}: We also report task-wise results on Office-Home and Office-31 in Figure \ref{fig:rfa_robustness_office_home} and Figure \ref{fig:rfa_robustness_office31}, respectively. 
We compare RFA with MDD Baseline (adopting normally pre-trained ImageNet model with MDD algorithm) and Robust PT (adopting adversarially pre-trained ImageNet model with MDD algorithm) with backbone model ResNet-50. It can be seen that our method improves the robustness significantly on all tasks.
The figures show a comparison of robustness and accuracy with three differently trained feature extractor student models. 






\begin{table*}
\centering
\begin{tabular}{llllllll}
    \toprule
    Dataset    & Robust PT                   & Source-only   & \multicolumn{1}{c}{DANN~\cite{ganin2016domain}}             & \multicolumn{1}{c}{DAN~\cite{long2015learning}}             & \multicolumn{1}{c}{CDAN~\cite{long2018conditional}}           & \multicolumn{1}{c}{JAN~\cite{long2017deep}}            & \multicolumn{1}{c}{MDD~\cite{zhang2019bridging}}           \\
    \midrule
    \multirow{6}{*}{VisDA-2017} & $\ell_{2}( \epsilon=0)$          & 43.05 / 0     & 71.34 / 0        & 61.79 / 0.01    & 74.23 / 0      & 63.70 / 0      & 72.20 / 4.03   \\
               & $\ell_{2}( \epsilon=3)$        & 30.57 / 4.20  & 66.01 / 36.57    & 48.54 / 18.40   & 67.52 / 39.78  & 56.18 / 29.41  & 59.79 / 35.38   \\
               & $\ell_{2}( \epsilon=5)$        & 31.24 / 5.27  & 63.95 / 35.50    & 44.87 / 18.16   & 67.43 / 41.20  & 54.14 / 29.58  & 59.80 / 35.35    \\
    \cline{2-8}
               & $\ell_{\infty}( \epsilon=2)$   & 37.55 / 3.58  & 70.30 / 36.56    & 54.51 / 18.23   & 71.48 / 38.59  & 59.13 / 28.55  & 67.72 / 39.50  \\
               & $\ell_{\infty}( \epsilon=4)$   & 34.47 / 4.72  & 65.79 / 38.21    & 48.65 / 19.99   & 68.00 / 41.67  & 55.08 / 32.15  & 60.97 / 37.46  \\
               & $\ell_{\infty}( \epsilon=8)$   & 25.67 / 6.64  & 63.45 / 37.44    & 42.24 / 22.11   & 65.18 / 41.67  & 52.00 / 31.87  & 52.78 / 32.06  \\
    \hline
    \multirow{6}{*}{Office-31}  & $\ell_2( \epsilon=0)$          & 77.80 / 0.02    & 85.79 / 0        & 81.72 / 0       & 86.90 / 0      & 85.68 / 0      & 88.31 / 1.70   \\
               & $\ell_{2}( \epsilon=3)$        & 68.61 / 34.05   & 77.78 / 55.06    & 73.14 / 33.99   & 78.93 / 57.49  & 78.20 / 50.12  & 80.00 / 61.21    \\
               & $\ell_{2}( \epsilon=5)$        & 64.08 / 30.55   & 73.70 / 55.39    & 69.34 / 34.05   & 74.75 / 58.43  & 73.38 / 49.03  & 75.72 / 60.87    \\
     \cline{2-8}
               & $\ell_{\infty}( \epsilon=2)$   & 73.91 / 35.57   & 81.58 / 58.70    & 77.81 / 38.09   & 82.43 / 60.41  & 81.93 / 52.61  & 84.05 / 64.62    \\
               & $\ell_{\infty}( \epsilon=4)$   & 69.51 / 41.11   & 77.30 / 62.38    & 73.71 / 42.29   & 79.67 / 65.53  & 78.88 / 57.85  & 80.72 / 67.54    \\
               & $\ell_{\infty}( \epsilon=8)$   & 65.62 / 39.54   & 74.24 / 61.73    & 70.61 / 40.40   & 75.65 / 64.72  & 75.12 / 60.24  & 75.73 / 66.46    \\
    \hline
        \multirow{6}{*}{Office-Home} & $\ell_2( \epsilon=0)$          & 58.29 / 0.06    & 63.39 / 0.05     & 59.64 / 0.23    & 67.03 / 0.04   & 64.61 / 0.07   & 67.91 / 5.81   \\
                & $\ell_{2}( \epsilon=3)$        & 51.45 / 24.03   & 56.82 / 30.39    & 52.98 / 18.45   & 61.08 / 35.77  & 58.84 / 24.92  & 62.04 / 38.06    \\
                & $\ell_{2}( \epsilon=5)$        & 48.85 / 21.32   & 53.67 / 28.33    & 50.70 / 17.40   & 58.10 / 33.34  & 56.43 / 24.20  & 59.24 / 36.62    \\
    \cline{2-8}
                & $\ell_{\infty}( \epsilon=2)$   & 56.02 / 27.74   & 60.76 / 34.44    & 57.43 / 21.47   & 65.08 / 41.15  & 63.37 / 29.65  & 65.85 / 42.93    \\
                & $\ell_{\infty}( \epsilon=4)$   & 53.89 / 31.46   & 58.10 / 37.25    & 55.18 / 24.21   & 63.04 / 43.81  & 60.74 / 33.09  & 63.30 / 43.42    \\
                & $\ell_{\infty}( \epsilon=8)$   & 49.87 / 28.89   & 54.79 / 36.01    & 51.48 / 23.20   & 59.10 / 42.80  & 57.10 / 32.99  & 59.56 / 42.66    \\
    \bottomrule
    
\end{tabular}
\caption{Effect of robust pre-training with varying perturbation budget ($\epsilon$) on unsupervised domain adaptation. Reported results are shown as \textbf{clean accuracy / adversarial robustness (\%)}.}
\label{tab:pretraining_extended}
\end{table*}

\end{appendices}

\end{document}